\documentclass[letterpaper]{article} % DO NOT CHANGE THIS
\usepackage[preprint]{aaai2027}  % arXiv preprint: show authors, no copyright slug
\usepackage[hyphens]{url}
\usepackage{graphicx}
\usepackage{natbib}
\usepackage{caption}
\usepackage{subcaption}
\usepackage{algorithm}
\usepackage{algorithmic}
\usepackage{newfloat}
\usepackage{listings}
\usepackage{booktabs}
\usepackage{amsmath,amssymb,amsthm}
\usepackage{tikz}
\usetikzlibrary{positioning,arrows.meta,shapes.geometric,decorations.pathreplacing,calc}

\newtheorem{theorem}{Theorem}

\newtheorem*{theorem*}{Theorem}
\newtheorem*{proposition*}{Proposition}

\newcommand{\bank}{\mathcal{B}}
\newcommand{\skset}{\mathcal{S}}
\newcommand{\uSMCO}{u\textsc{-SMCO}}
\newcommand{\CASS}{\textsc{CASS}}

\usepackage[most]{tcolorbox}
\usepackage{colortbl}
\definecolor{figgreen}{HTML}{2E8B57}
\definecolor{figpurple}{HTML}{7A4FA3}
\definecolor{casered}{HTML}{C0504D}
\definecolor{casegreen}{HTML}{2E8B57}
\newtcolorbox{skillbox}[2]{enhanced,colback=white,colframe=#1,boxrule=0.6pt,arc=2pt,
  left=4pt,right=4pt,top=2pt,bottom=2pt,boxsep=0pt,before skip=1.5pt,after skip=1.5pt,
  fonttitle=\scriptsize\bfseries,coltitle=white,colbacktitle=#1,
  title=#2,attach boxed title to top left={xshift=4pt,yshift=-2pt},
  boxed title style={boxrule=0pt,arc=1pt,left=3pt,right=3pt,top=1pt,bottom=1pt}}
\newtcolorbox{findingbox}{
    enhanced,
    colback=white,
    colframe=black!60,
    boxrule=0.7pt,
    arc=2.5pt,
    left=5pt,
    right=5pt,
    top=3pt,
    bottom=3pt,
    boxsep=0pt,
    before skip=4pt,
    after skip=4pt,
    fontupper=\bfseries\itshape
}

\title{Coalition-Aware Skill Reliability for Self-Evolving Agents}

\author{
    Qiyan Zhao\textsuperscript{\rm 1,2},
    Xiaofeng Zhang\textsuperscript{\rm 2,3},
    Bo Liu\textsuperscript{\rm 1},
    Minda Chen\textsuperscript{\rm 2},
    Wei Xiong\textsuperscript{\rm 2},\\
    Jingyang Chen\textsuperscript{\rm 2},
    Guanting Ye\textsuperscript{\rm 5},
    Wenhao Yu\textsuperscript{\rm 5},
    Xiaosong Yuan\textsuperscript{\rm 3},
    Shijie Han\textsuperscript{\rm 4},\\
    Da-Han Wang\textsuperscript{\rm 1},
    Jianmin Ji\textsuperscript{\rm 5},
    Fei Huang\textsuperscript{\rm 2},
    Xu-Yao Zhang\textsuperscript{\rm 1,$\dagger$}
}

\affiliations{
    \textsuperscript{\rm 1}CASIA\quad
    \textsuperscript{\rm 2}LongShine AI Lab\quad
    \textsuperscript{\rm 3}SJTU\quad
    \textsuperscript{\rm 4}HKUST\quad
    \textsuperscript{\rm 5}USTC\\
    \textsuperscript{$\dagger$}Corresponding author
}

\begin{document}

\maketitle

\begin{abstract}
Agent skills, structured artifacts distilled from interaction trajectories and dynamically reused from skill banks, have become a central mechanism for enabling large language model (LLM)-based self-evolving agents to learn from past experience. Yet existing work has largely focused on the operational aspects of skills, such as acquisition, evolution, and retrieval, while leaving a more fundamental reliability question unresolved: \emph{Do accumulated skills in an agent's skill bank actually make positive mechanistic contributions?} We investigate this question through systematic skill-bank audits across alternative bank compositions and deployment domains, measuring the resulting changes in agent behavior. These audits reveal two recurring reliability failures: \textbf{coalition pollution}, where bank-level gains conceal negative coalition-level skill contributions, and \textbf{cross-domain utility reversal}, where source-beneficial skills reverse their effects after transfer. These findings motivate two reliability interventions: coalition-aware skill selection during skill accumulation and label-free skill masking after transfer. \textbf{Coalition-Aware Skill Selection (CASS)} selects more reliable candidate skills for the current bank using sampled Shapley marginals. \textbf{Unsupervised Skill-Masked Coalition Optimizer (u-SMCO)} masks transferred skills whose exclusion improves retrieval quality on unlabeled target-domain data. Agentic experiments on LoCoMo, LongMemEval, HotpotQA, and ALFWorld show that CASS and u-SMCO consistently improve task performance and cross-domain generalization over strong skill-based self-evolving agent baselines. Beyond accuracy, coalition-conditioned reliability modeling reduces sensitivity to noisy outcome-reward fluctuations during reinforcement learning and exposes the limits of isolation-based skill evaluation. These results establish skill reliability as a joint property of skills, banks, and deployment domains, with skill coalitions serving as the central unit of reliable skill evolution. Our audit toolkit, method code, and trained checkpoints will be released.
\end{abstract}

\begin{figure}[t]
    \centering
    \vspace{-0.2mm}
    \includegraphics[width=1\linewidth]{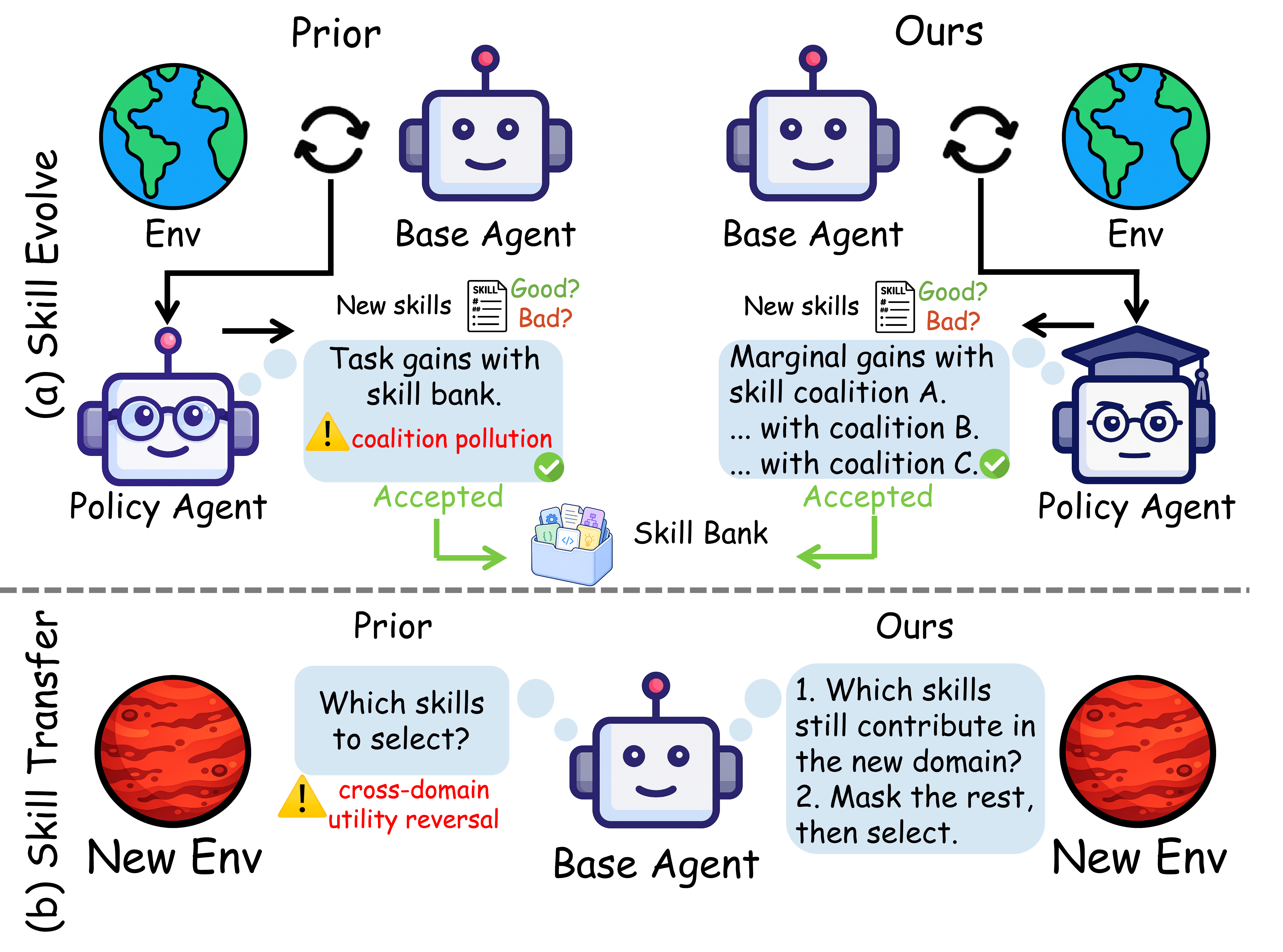}
    \vspace{-6.3mm}
    \caption{\textbf{Motivation.} (a) Existing self-evolving agents overlook coalition pollution during skill evolution, admitting unreliable skills with negative coalition-level contributions. (b) During cross-domain transfer, prior methods ignore cross-domain utility reversal, allowing skills that become unreliable in the target domain to influence subsequent decisions. These two mechanistic failure modes motivate our work.
}
    \vspace{-0.4mm}
    \label{motivation-fig1}
\end{figure}

\section{Introduction}

Large language model (LLM)-based agents~\cite{wei2026agentic, huang2026raw, wang2023voyager} are increasingly capable of self-evolving by learning from both successful and failed interactions. Recent works~\cite{xia2026skillrl, zhang2026memskill, yang2026skillopt, shi2026skill1, ouyang2026skillos} advance this paradigm by distilling raw interaction trajectories into \emph{skills}~\cite{xu2026agent}: high-level, reusable, structured artifacts that abstract the essential knowledge for solving tasks. Once accumulated into a persistent skill bank, these artifacts are retrieved and instantiated at inference time without additional parameter updates~\cite{berthon2026skill, li2026skillsbench, xu2026skill}.

\begin{figure*}[t]
    \centering
    \vspace{-0.9em}
    \includegraphics[width=0.98\textwidth]{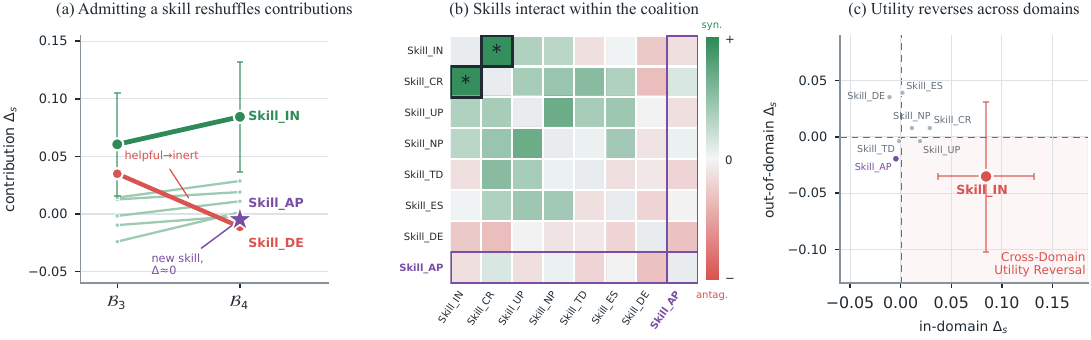}
    \vspace{-0.3em}
    \caption{\textbf{Skill reliability is not intrinsic.} We audit the same skills across bank compositions and deployment domains (skills relabeled \texttt{Skill\_XX}). (a) \emph{Acceptance is not contribution:} the gate admits \texttt{Skill\_AP} (purple star), yet it contributes ${\approx}0$ while the incumbent \texttt{Skill\_IN} carries the bank. (b) \emph{Contribution depends on partners:} the interaction matrix reveals synergies and antagonisms, strongest between \texttt{Skill\_IN} and \texttt{Skill\_CR} (boxed). (c) \emph{Same bank, different verdict:} \texttt{Skill\_IN} helps at the source (LoCoMo) and reverses at the target (HotpotQA-50).}
    \label{analysis-fig2}
    \vspace{-1.6em}
\end{figure*}

Most existing work focuses on operations along the skill lifecycle, particularly skill acquisition~\cite{qiu2026autorefine, ni2026trace2skill}, evolution~\cite{zhang2026coevoskills, shen2026skillfoundry}, and retrieval~\cite{xia2026skillrl, su2026skill}. Other studies have investigated how individual skills can be composed~\cite{chen2026skillcraft, li2026skillgraph} and evaluated their generalization across tasks and domains~\cite{huang2026raw, li2026skillsbench}. Despite these practical successes in expanding agent capabilities, a more fundamental reliability question remains largely unexplored: \textbf{\emph{Do the skills accumulated in an agent's skill bank actually make positive mechanistic contributions when invoked?}}

To this end, we conduct a systematic Skill Mechanistic Reliability Audit (SMRA) on representative skill-based self-evolving agents~\cite{zhang2026memskill}. Our audit covers two lifecycle stages: skill evolution and cross-domain transfer (Figure~\ref{motivation-fig1}). Both stages audit skill reliability by systematically varying the skill-bank composition. The evolution stage compares bank states before and after new skills are accepted (Figure~\ref{analysis-fig2}a-b); the transfer stage compares in-domain versus out-of-domain deployment on the accumulated bank (Figure~\ref{analysis-fig2}c). This audit reveals two key reliability failures:

\noindent
\textbf{Coalition Pollution.} During skill evolution, a newly extracted candidate is accepted into the skill bank if its inclusion improves aggregate task performance. Yet this gain reveals neither which skill produces the value nor how skills interact within the bank. Comparing the audit results before and after acceptance (Figure~\ref{analysis-fig2}a-b), we uncover an inversion: the admitted skill shows near-zero contribution while an incumbent carries the bank, and existing marginals shift with the composition. The gate thus retains skills whose mechanistic reliability it never established, because a skill's contribution is conditioned on its coalition rather than intrinsic. We term this phenomenon coalition pollution.

\noindent
\textbf{Cross-Domain Utility Reversal.} During cross-domain transfer, agents carry the entire accumulated skill bank into the target domain and dynamically retrieve skills for downstream tasks. This practice implicitly assumes that skills useful in the source domain remain useful after transfer. Comparing the audit results across in-domain and out-of-domain deployment (Figure~\ref{analysis-fig2}c), we find this assumption fails: some skills beneficial in the source domain reverse their effect at the target and degrade performance. We term this failure mode cross-domain utility reversal.

\begin{findingbox}
Self-evolving agents' skill banks contain two reliability failure modes: bank-level gains conceal coalition-level harm during evolution, and source-useful skills reverse to harmful under transfer.
\end{findingbox}

Based on these findings, we introduce two interventions, one for each failure mode.

\textbf{Coalition-Aware Skill Selection (CASS)} addresses coalition pollution during skill evolution. Instead of relying on aggregate reward alone, CASS augments the acceptance criterion with a coalition-aware signal. When evaluating a candidate, it uses Monte Carlo sampling to construct diverse skill coalitions from the resulting bank and aggregates their Shapley marginal contributions to estimate coalition-conditioned reliability. This estimate is then combined with aggregate reward to decide acceptance.

\textbf{Unsupervised Skill-Masked Coalition Optimizer (u-SMCO)} addresses cross-domain utility reversal after transfer. Using unlabeled queries from the target domain, u-SMCO scores each skill by its retrieval-quality contribution and greedily masks skills whose removal improves retrieval. Because no task labels are required, u-SMCO applies at deployment without target-domain supervision.

Experiments on LoCoMo, LongMemEval, HotpotQA, and ALFWorld show that CASS and u-SMCO consistently improve task performance and cross-domain generalization. Further analysis shows that our coalition-aware signals mitigate reward noise during training; we further prove that outcome-only gates are structurally blind to the coalition-level interactions these failures rest on. Together, these results reveal a broader principle: skill reliability is a joint property of the skill, the surrounding skill bank, and the deployment domain, not an intrinsic attribute of individual skills.

Our contributions can be summarized as follows.

\begin{itemize}\itemsep=0pt

\item We introduce SMRA and identify two reliability failure modes of self-evolving skills: coalition pollution during evolution and cross-domain utility reversal after transfer.

\item We propose CASS, a coalition-aware method that selects more reliable candidate skills for the bank by augmenting aggregate reward with sampled Shapley marginals.

\item We propose u-SMCO, a label-free method that masks skills whose utility reverses after transfer using only unlabeled target-domain queries.

\item We prove that outcome-only gates are structurally blind to the coalition-level interactions that drive both failure modes, motivating our two interventions.

\item Across four benchmarks, CASS and u-SMCO consistently improve task performance and cross-domain generalization over strong self-evolving baselines.

\end{itemize}

\section{Related Work}

\subsection{Self-Evolving LLM Agents}

Large language models (LLMs)~\cite{huang2024llama, yang2025qwen3, agarwal2025gpt} have demonstrated strong reasoning capabilities across a wide range of complex tasks~\cite{xu2026skill, maharana2024evaluating, wu2025longmemeval, yang2018hotpotqa, shridhar2020alfworld}, accelerating the development of autonomous agent systems~\cite{wei2026agentic, wang2023voyager}. Some studies have further explored in-context learning to enhance agent reasoning~\cite{wu2023autogen, zhao2024expel}. However, these agents typically treat each interaction as an isolated episode and approach every new task from scratch, without leveraging prior experience. This paradigm limits their ability to adapt to increasingly complex or long-horizon tasks in dynamic, open-ended environments.

To enable LLM-based agents to learn from past interactions, memory-based self-evolving agents~\cite{xu2026mem, wang2025mirix, yu2024chain, kang2025memory} store sampled interaction trajectories in external databases as reusable experience. However, raw trajectories often contain substantial redundancy and noise~\cite{zhao2024expel, chhikara2025mem0, mi2026procmem}, motivating the development of skill-based self-evolving agents~\cite{zhang2025memgen, ouyang2025reasoningbank, wei2025evo} that distill historical trajectories into compact, reusable behavioral primitives, namely skills~\cite{tu2026dynamic, zhu2026skill0}. In parallel, advances in reinforcement learning~\cite{chen2026learning, yu2026dapo, wang2026reinforcement} have provided stronger supervisory signals for agent self-evolution.

\subsection{Agent Skills}

With the emergence of agent scaffolds such as Claude Code and OpenClaw~\cite{xu2026agent, li2026organizing, he2026openclaw}, \emph{Agent Skills} have become a structured mechanism for encoding reusable decision-making strategies~\cite{shen2026dynamic, liang2026skillnet}. These behavioral primitives can be retrieved at inference time without updating model parameters, allowing agents to reuse knowledge distilled from both successful and failed interaction trajectories~\cite{qiu2026autorefine, mi2026procmem}. 

Existing work spans the full skill lifecycle~\cite{xu2026agent, vishe2026skill}: Trace2Skill~\cite{ni2026trace2skill} extracts skills from execution logs, MemP~\cite{fang2026memp} formalizes construction, retrieval, and updating as executable programs, MemSkill~\cite{zhang2026memskill} develops skills for memory operations, CoEvoSkills~\cite{zhang2026coevoskills} refines skills through co-evolutionary validation, SkillRL~\cite{xia2026skillrl} and other reinforcement-learning approaches~\cite{alzubi2026evoskill, wang2026reinforcement} refine skill banks with reward signals, Skill0~\cite{lu2026skill0} internalizes acquired skills into model parameters, and SkillGraph~\cite{li2026skillgraph} models inter-skill dependencies with graph structures. Despite these advances, existing methods largely emphasize the operational aspects of skills, leaving a fundamental reliability question unresolved: \emph{Do accumulated skills in an agent's skill bank make positive mechanistic contributions when invoked?} Our work addresses this question by examining skill reliability during evolution and cross-domain transfer.

\subsection{Skill Reliability}
\label{ssec:rw-reliability}

The concurrent work most closely related to ours is SkillLens~\citep{huang2026raw}, which characterizes skill reliability from a utility-grounded perspective and shows that textually plausible skills can still have negative effects. Our work is complementary along three axes: we examine reliability through \emph{causal} contributions to downstream performance rather than text-based utility estimation; we adopt a \emph{coalition-aware} view in which reliability is a joint property of the skill, its surrounding bank, and the deployment domain, rather than an intrinsic per-skill attribute; and we go beyond diagnosis to propose two mechanistic interventions that address the failure modes we identify.

Our approach builds on two methodological lineages: coalition-level contribution attribution via the Shapley value~\citep{shapley1953value, lundberg2017unified} and its higher-order interaction extensions~\citep{sundararajan2020many, grabisch2016set}; and model reliability under distribution shift~\citep{iwasawa2021test, wang2020tent}, which shows that source-domain competence is a poor predictor of target-domain behavior.

\section{Skill Mechanistic Reliability Audit}
\label{sec:motivation}

In this section, we begin with a brief overview of the baseline agent's skill-bank evolution mechanism. We then apply SMRA to trace skill contributions across bank compositions and deployment domains, revealing coalition pollution and cross-domain utility reversal. Finally, we analyze why these failures arise, motivating our method design.

\begin{figure*}[t]
    \centering
    \includegraphics[width=\textwidth]{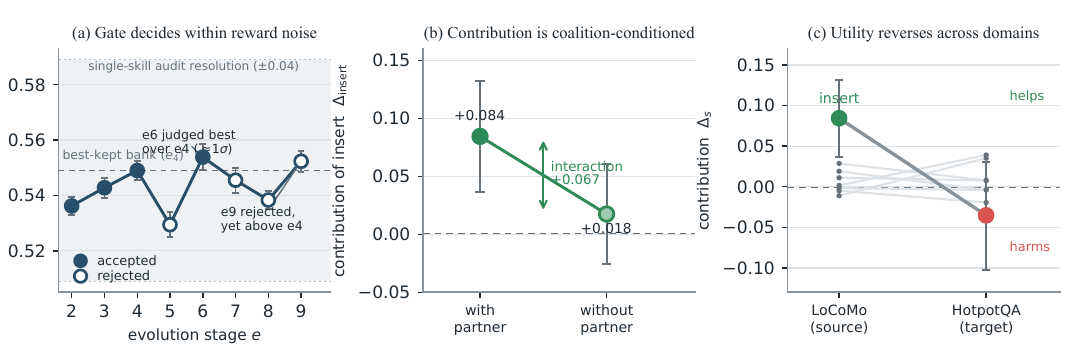}
    \vspace{-2.0em}

    \caption{\textbf{Reliability failures revealed by SMRA.} (a) \emph{Decisions at the noise floor:} accept/reject margins are commensurate with the $\pm 1$~SE error bars; the shaded band marks the $\pm 0.04$ single-skill audit resolution, an order of magnitude wider. (b) \emph{Partner removed, marginal drops:} \texttt{insert} falls from $+0.084$ to $+0.018$ once \texttt{capture\_contextual\_relationships} is removed. (c) \emph{Source benefit, target reversal:} \texttt{insert} helps on the source (LoCoMo), reverses on the target (HP-50).}
        \vspace{-1.5em}

    \label{fig:reliability-trajectories}
\end{figure*}

\subsection{Self-Evolution and SMRA Formulation}

Consider a skill-based agent built on a base agent $F$. At evolution stage $e$, let $\mathcal{B}_e=\{s_1,\ldots,s_n\}$ denote its current skill bank, where each $s_i$ is a reusable skill. Given the interaction history accumulated by $F$, a skill-evolution policy $F'$ proposes a candidate skill $s_{\mathrm{new}}$, yielding the candidate bank $\widetilde{\mathcal{B}}_{e+1}=\mathcal{B}_e\cup\{s_{\mathrm{new}}\}$. Under an outcome-only gate, whether $s_{\mathrm{new}}$ is retained is determined by the bank-level gain
\begin{equation}
G_e =
r_{\mathrm{out}}(\widetilde{\mathcal{B}}_{e+1})
-
r_{\mathrm{out}}(\mathcal{B}_e),
\label{eq:outcome-gain}
\end{equation}
where $r_{\mathrm{out}}(\mathcal{B})$ is the aggregate outcome reward estimated online from recent training interactions with bank $\mathcal{B}$. If $G_e>0$, the candidate is accepted and $\mathcal{B}_{e+1}=\widetilde{\mathcal{B}}_{e+1}$; otherwise, $\mathcal{B}_{e+1}=\mathcal{B}_e$. This gate captures only bank-level gains, not how skill contributions vary across bank compositions.

To test whether accumulated skills make positive mechanistic contributions, we introduce the \emph{Skill Mechanistic Reliability Audit} (SMRA). SMRA adopts Shapley's marginal-contribution view~\citep{shapley1953value,lundberg2017unified}: skills are players, and $V^D(\mathcal{C})$ gives the downstream value of any skill coalition $\mathcal{C}\subseteq\mathcal{B}$ on domain $D$. For each $s\in\mathcal{B}$, SMRA audits its full-bank marginal contribution through a knockout intervention:
\begin{equation}
\Delta_s^D(\mathcal{B})
=
V^D(\mathcal{B})
-
V^D(\mathcal{B}\setminus\{s\}).
\label{eq:knockout-contribution}
\end{equation}
Full-bank and knockout evaluations are paired on identical examples. Positive $\Delta_s^D$ means $s$ supports the bank in $D$; negative values mean its knockout improves performance. To account for composition dependence, the Shapley value generalizes this full-bank marginal by taking a weighted average over every subset $\mathcal{C}\subseteq\mathcal{B}\setminus\{s\}$ of the other skills:
\begin{equation}
\begin{aligned}
\phi_s^D(\mathcal{B})
&=
\sum_{\mathcal{C}\subseteq\mathcal{B}\setminus\{s\}}
\frac{|\mathcal{C}|!(|\mathcal{B}|-|\mathcal{C}|-1)!}
     {|\mathcal{B}|!} \\
&\quad\cdot
\left[
V^D(\mathcal{C}\cup\{s\})
-
V^D(\mathcal{C})
\right].
\end{aligned}
\label{eq:shapley-contribution}
\end{equation}
Thus, $\phi_s^D(\mathcal{B})$ summarizes the expected marginal contribution of $s$ across alternative bank compositions. Collectively, these values form a mechanistic reliability profile of $\mathcal{B}$. Repeating the audit across evolving banks and deployment domains exposes coalition pollution and cross-domain utility reversal, respectively.

\subsection{Coalition Pollution}
\label{ssec:typeA}

An outcome-only gate judges a candidate by a single scalar: it accepts when the bank's aggregate reward rises, $G_e>0$. That scalar is an online, bank-level summary of the transition; it does not measure which skill in the resulting bank contributes. Our audit asks that separate question, isolating each skill by a knockout paired on identical held-out examples (Eq.~\ref{eq:knockout-contribution}). In our baseline agent, the gate admits $\mathcal{B}_3\!\to\!\mathcal{B}_4$ on an online margin of only $+0.006$, which it cannot ascribe to any single skill; the paired audit of the resulting bank (Table~\ref{tab:audit}) then finds the newly admitted \texttt{capture\_activity\_preferences} non-contributing ($\Delta=-0.005$), its pairwise interactions likewise negligible (Figure~\ref{analysis-fig2}b), while \texttt{insert} is the skill actually carrying the bank ($\Delta=+0.084$). Nor is contribution an intrinsic property of a skill: the marginal of \texttt{insert} drops sharply once its coalition partner \texttt{capture\_contextual\_relationships} is removed (Figure~\ref{fig:reliability-trajectories}(b)). Reliability is therefore a property of the coalition rather than the individual skill; we term the retention of non-contributing skills under such a gate \emph{coalition pollution}.

\begin{table}[t]
\centering
\scriptsize
\setlength{\tabcolsep}{1pt}
\captionof{table}{\textbf{Per-skill SMRA contributions} before ($\mathcal{B}_3$) and after ($\mathcal{B}_4$) the gate's acceptance (LoCoMo). $^{*}$: $95\%$ CI excludes zero; $^{\dagger}$: newly accepted.}
\label{tab:audit}
\vspace{-2.5pt}
\begin{tabular*}{\columnwidth}{@{\extracolsep{\fill}} l r r l @{}}
\toprule
Skill & $\Delta_s(\mathcal{B}_3)$ & $\Delta_s(\mathcal{B}_4)$ & Full name \\
\midrule
\texttt{Skill\_IN} & $+0.061^{*}$ & $+0.084^{*}$ & \texttt{insert} \\
\texttt{Skill\_DE} & $+0.035$ & $-0.011$ & \texttt{delete} \\
\texttt{Skill\_UP} & $+0.013$ & $+0.019$ & \texttt{update} \\
\texttt{Skill\_NP} & $-0.002$ & $+0.011$ & \texttt{noop} \\
\texttt{Skill\_TD} & $-0.010$ & $-0.002$ & \texttt{capture\_temporal\_details} \\
\texttt{Skill\_CR} & $+0.014$ & $+0.029$ & \texttt{capture\_contextual\_relationships} \\
\texttt{Skill\_ES} & $-0.024$ & $+0.002$ & \texttt{capture\_entity\_specifics} \\
\texttt{Skill\_AP}$^{\dagger}$ & --- & $-0.005$ & \texttt{capture\_activity\_preferences} \\
\bottomrule
\end{tabular*}
\end{table}

The gate cannot prevent this. Figure~\ref{fig:reliability-trajectories}(a) shows why: the accept/reject margins it acts on ($0.006$ to $0.02$) are only one to a few times the per-stage standard error of its own reward estimate (error bars, $\approx 0.004$), so statistically indistinguishable banks are ranked as decisively better or worse. The per-skill contributions that would justify these rankings are further out of reach: auditing a single skill on $314$ paired queries resolves its contribution only to $\approx\pm 0.04$ (shaded band), an order of magnitude coarser than the margins being acted on. This is not merely a sampling limitation but a structural one.

\begin{theorem}[Outcome-only non-identifiability]
\label{thm:existence}
For a gate observing only the endpoints $V(\mathcal{B}_e)$ and $V(\widetilde{\mathcal{B}}_{e+1})$, a positive gain $V(\widetilde{\mathcal{B}}_{e+1})-V(\mathcal{B}_e)>0$ does not identify whether $\widetilde{\mathcal{B}}_{e+1}$ contains a negatively contributing skill.
\end{theorem}

In the audit's terms, a noise-free gate reads exactly one entry of the bank's reliability profile: $G_e$ is the candidate's own marginal $\Delta_{s_{\mathrm{new}}}(\widetilde{\mathcal{B}}_{e+1})$; every other member is left unconstrained, and Appendix~A.1 constructs two banks with identical endpoints yet opposite signs for one member's contribution. \CASS{} closes this gap by sampling coalition knockouts during evolution, an audit under alternative compositions.

\vspace{-1pt}
\subsection{Cross-Domain Utility Reversal}
\label{ssec:typeB}

A skill selected in one domain is carried unchanged into another, implicitly assuming its usefulness transfers. It need not. Holding the bank fixed and varying only the deployment domain, a skill exhibits \emph{cross-domain utility reversal} when its contribution is positive at the source $D_{\mathrm{src}}$ yet negative at the target $D_{\mathrm{tgt}}$,
\begin{equation}
\Delta_s^{D_{\mathrm{src}}}(\mathcal{B})>0
\quad\text{and}\quad
\Delta_s^{D_{\mathrm{tgt}}}(\mathcal{B})<0 .
\label{eq:domain-flip}
\end{equation}
The clearest case is \texttt{insert}. On the source domain LoCoMo it is significantly beneficial ($\Delta=+0.084$, $95\%$ CI excluding zero), yet this benefit does not survive transfer: on HotpotQA its point estimate reverses to $\Delta=-0.035$, though with a $95\%$ CI that still includes zero (Figure~\ref{fig:reliability-trajectories}(c)). We therefore read the target side as a directional reversal rather than significant harm. Because the bank is identical in both evaluations and only the deployment distribution changes, a contribution that is reliably positive at the source carries no such guarantee at the target.

The cause is that the value function is itself domain-dependent: the target distribution changes both where a skill is exercised and what its contributions are worth, so a source-domain contribution cannot be carried across domains as an invariant. A training-time gate sees only $V^{D_{\mathrm{src}}}$ and has no signal fixing the sign of $\Delta_s^{D_{\mathrm{tgt}}}$. Correcting cross-domain utility reversal therefore requires observing the target distribution; because target labels are often unavailable, \uSMCO{} relies instead on an unlabeled retrieval-quality signal, masking skills whose removal improves retrieval on target queries.

\subsection{A Unified Reliability View}

Coalition pollution and cross-domain utility reversal are two faces of one fact: skill reliability is not intrinsic. A skill's contribution depends on the bank it sits in and on the domain where it is deployed, so we write it as a contextual quantity $R(s;\mathcal{B},D)$ rather than a fixed score $R(s)$. An outcome-only gate, which reads a single bank-level scalar at the source domain, is blind to both dependencies by construction. The two failures also mark where the missing signal can be restored: during evolution, \CASS{} re-audits the candidate bank under sampled coalition knockouts before retaining it, supplying the composition dependence the gate lacks; after transfer, \uSMCO{} re-scores the transferred bank with a label-free, target-conditioned retrieval signal, supplying the domain dependence the gate never observed. Both apply the same principle at the two stages where context changes: reliability is judged on coalitions, never on skills in isolation.

\begin{findingbox}
Do accumulated skills actually contribute? The question has no skill-intrinsic answer: contribution depends on the surrounding coalition and the deployment domain, and bank-level success is evidence of neither.
\end{findingbox}

%======================================================================
\section{Method}
\label{sec:method}
%======================================================================

To address coalition pollution and cross-domain utility reversal, we propose two reliability interventions, Coalition-Aware Skill Selection (\CASS{}) and Unsupervised Skill-Masked Coalition Optimizer (\uSMCO{}), respectively.

\subsection{CASS: Coalition-Aware Skill Selection}
\label{ssec:cass}

\begin{figure}[t]
\centering
\includegraphics[width=\linewidth]{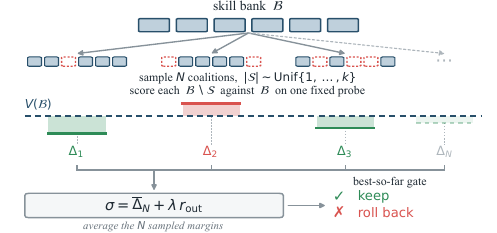}
% The figure PDF is cropped flush to its ink (build_fig4.py), so this skip is
% the whole gap between drawing and caption; 4pt keeps them visually one block.
% Note this cannot buy space for the body: aaai2027.sty sets \flushbottom, so
% every column is stretched to the bottom and any slack returns as paragraph
% glue. Only cutting words shortens the paper.
\setlength{\abovecaptionskip}{4pt}
\caption{\textbf{One \CASS{} gate decision.} Each sampled coalition is knocked out and scored against the full bank on one fixed probe, so every margin is a paired difference from $V(\bank)$; green marks removal that costs value, red that gains it. Algorithm~A1 gives the full procedure.}
\vspace{-1pt}
\label{fig:cass}
\end{figure}

Section~\ref{ssec:typeA} audited reliability after the fact; \CASS{} embeds that audit into skill evolution itself. Whenever the gate chooses between candidate banks that differ in composition, the signal it needs is the composition-averaged contribution formalized by Eq.~\eqref{eq:shapley-contribution}. Estimating that value to useful precision takes hundreds of permutation samples per skill (Appendix~A.3), each rebuilding the agent's memory under the ablated composition and re-evaluating the bank, which is prohibitive at every gate decision. \CASS{} therefore retains the coalition view but abandons per-skill attribution, estimating instead a bank-level coalition average by Monte Carlo sampling over a bounded number of coalitions, trading exhaustive evaluation for a fixed stochastic budget as in sampling-based planning~\citep{kocsis2006bandit, grabisch2016set}.

Given a bank $\bank$ and a coalition $\skset\subseteq\bank$, a subset of the skills it holds, \CASS{} scores the bank in two steps. It first draws $N$ distinct coalitions by Monte Carlo sampling over the bank (Figure~\ref{fig:cass}): each draw takes a size $|\skset|$ uniform on $\{1,\dots,k\}$, then a coalition $\skset$ uniform among the subsets of that size. It then evaluates every sampled coalition by its knockout margin,
\begin{equation}
\Delta_{\skset}(\bank)=\frac{V(\bank)-V(\bank\setminus\skset)}{|\skset|},
\label{eq:coalition-margin}
\end{equation}
which charges the joint loss of removing $\skset$ equally to its members. Sizes above one are what make the score more than a per-skill average: the joint loss departs from the sum of single-skill margins exactly when skills interact. \CASS{} averages the $N$ margins into $\widehat{\bar\Delta}_N(\bank)$ and combines it with the online outcome reward into the scalar the gate ranks,
\begin{table*}[t]
\centering
\small
% 普通 tabular + 加宽 tabcolsep 撑满栏宽：tabular* 的 \extracolsep{\fill}
% 胶不会被 \rowcolor 染色，末两行的灰底会断成一格一格
\setlength{\tabcolsep}{7.2pt}
\renewcommand{\arraystretch}{1.15}
\begin{tabular}{l cc cc cc cc cc c}
\toprule
 & \multicolumn{6}{c}{\textbf{Conversational Benchmarks}} & \multicolumn{5}{c}{\textbf{Embodied Interactive Tasks}} \\
\cmidrule(lr){2-7}\cmidrule(lr){8-12}
 & \multicolumn{2}{c}{\textbf{LoCoMo}} & \multicolumn{2}{c}{\textbf{LongMemEval}} & \multicolumn{2}{c}{\textbf{Avg.}} & \multicolumn{2}{c}{\textbf{ALF-Seen}} & \multicolumn{2}{c}{\textbf{ALF-Unseen}} & \textbf{Avg.} \\
\cmidrule(lr){2-3}\cmidrule(lr){4-5}\cmidrule(lr){6-7}\cmidrule(lr){8-9}\cmidrule(lr){10-11}\cmidrule(lr){12-12}
Method & F1 & L-J & F1 & L-J & F1 & L-J & SR & \#Stps$\downarrow$ & SR & \#Stps$\downarrow$ & SR \\
\midrule
CoN & $30.86$ & $41.72$ & $30.78$ & $56.44$ & $30.82$ & $49.08$ & $75.00$ & $19.15$ & $80.60$ & $17.38$ & $77.80$ \\
ReadAgent & $28.63$ & $38.25$ & $24.48$ & $42.62$ & $26.56$ & $40.44$ & $62.86$ & $26.14$ & $71.64$ & $22.88$ & $67.25$ \\
MemoryBank & $36.80$ & $44.43$ & $30.56$ & $41.96$ & $33.68$ & $43.20$ & $60.71$ & $28.23$ & $66.42$ & $24.64$ & $63.57$ \\
A-MEM & $39.39$ & $49.71$ & $25.83$ & $38.04$ & $32.61$ & $43.88$ & $62.86$ & $27.53$ & $70.15$ & $23.79$ & $66.51$ \\
Mem0 & $25.48$ & $34.58$ & $30.25$ & $46.81$ & $27.87$ & $40.70$ & $74.29$ & $19.77$ & $81.34$ & $17.15$ & $77.82$ \\
MemoryOS & $41.39$ & $48.64$ & $17.59$ & $39.83$ & $29.49$ & $44.24$ & $57.86$ & $27.94$ & $65.67$ & $24.46$ & $61.77$ \\
\midrule
\textsc{MemSkill}$^{\dagger}$ & $39.45$ & $54.78$ & $30.02$ & $53.47$ & $34.74$ & $54.13$ & $74.01$ & $19.97$ & $80.09$ & $19.04$ & $77.05$ \\
\textbf{Ours} & $\mathbf{41.48}$ & $\mathbf{56.58}$ & $\mathbf{30.03}$ & $\mathbf{54.46}$ & $\mathbf{35.76}$ & $\mathbf{55.52}$ & $\mathbf{76.10}$ & $\mathbf{19.94}$ & $\mathbf{83.22}$ & $\mathbf{19.93}$ & $\mathbf{79.66}$ \\
\bottomrule
\end{tabular}
\caption{\textbf{Main results} (\%). L-J is the LLM-as-Judge score; \#Stps is the mean number of environment interactions (lower is better). $^{\dagger}$~our reproduction from \textsc{MemSkill}'s released project.}
\vspace{-8pt}
\label{tab:main}
\end{table*}
\begin{equation}
\sigma(\bank)=\widehat{\bar\Delta}_N(\bank)+\lambda\,r_{\mathrm{out}}(\bank),
\qquad\lambda=0.2,
\label{eq:cass-score}
\end{equation}
retaining $\widetilde{\bank}_{e+1}$ iff its score exceeds every score seen so far. The rule is the baseline's best-so-far gate of Eq.~\eqref{eq:outcome-gain} with only the ranked scalar changed.

\subsection{\uSMCO{}: Unsupervised Bank Masking}
\label{ssec:usmco}

\CASS{} handles coalition pollution \emph{during training}; cross-domain utility reversal instead calls for action at deployment. Given a trained bank $\bank$ and unlabeled target queries $Q_\text{tgt}$, \uSMCO{} decides which skills to mask, so it needs a per-skill signal that is computable without labels and causally connected to downstream utility. Label-free is the easy half. The agent's own preference, the drop in selection-policy entropy when a skill is present, ranks \texttt{insert} highest on HP-50, the very skill SMRA exposes as negative at the target (Figure~\ref{analysis-fig2}c): that policy was trained under the outcome gate and inherits its bias instead of tracking utility, echoing the collapse mode of entropy-minimization TTA \citep{iwasawa2021test}. Appendix~C.1 reports the ablation.

\uSMCO{} instead scores skills by retrieval quality on the target itself. For a set of skills $\mathcal{A}\subseteq\bank$, let $\mathcal{M}(\mathcal{A})$ be the memory bank rebuilt over the target contexts using only $\mathcal{A}$, and let $\text{Top}(q)$ collect the memories of $\mathcal{M}(\mathcal{A})$ nearest $q$ in cosine similarity. Define
\begin{equation}
\mathrm{RQ}(\mathcal{A})=\frac{1}{|Q_\text{tgt}|}\sum_{q\in Q_\text{tgt}}\ \text{mean}_{m\in\text{Top}(q)}\cos(e_q,e_m).
\label{eq:retrieval-signal}
\end{equation}
The mask score $\psi(s)=\mathrm{RQ}(\bank)-\mathrm{RQ}(\bank\setminus\{s\})$ is the audit's knockout comparison once more, scored by retrieval quality on the target rather than by task outcome. Unlike the entropy signal it is decoupled from the training reward channel: it asks whether a memory-writing skill improves the semantic locality of what it stores with respect to the queries the deployment distribution will ask.

\begin{algorithm}[t]
\caption{\uSMCO{}: Unsupervised Skill Masking}
\label{alg:usmco}
\begin{algorithmic}[1]
\REQUIRE bank $\bank$, unlabeled queries $Q_\text{tgt}$, contexts $C_\text{tgt}$, stop threshold $\tau$
\STATE $\text{kept} \leftarrow \bank$
\WHILE{$|\text{kept}| > 1$}
    \FOR{each $s \in \text{kept}$}
        \STATE Rebuild memory $\mathcal{M}(\text{kept}\!\setminus\!\{s\})$ over $C_\text{tgt}$
        \STATE $\psi(s) \leftarrow \mathrm{RQ}(\text{kept}) - \mathrm{RQ}(\text{kept}\!\setminus\!\{s\})$
    \ENDFOR
    \STATE $s^\star \leftarrow \arg\min_s \psi(s)$
    \IF{$\psi(s^\star) \geq \tau$}
        \STATE \textbf{break}
    \ELSE
        \STATE $\text{kept} \leftarrow \text{kept}\!\setminus\!\{s^\star\}$
    \ENDIF
\ENDWHILE
\RETURN $\text{kept}$
\end{algorithmic}
\end{algorithm}

Algorithm~\ref{alg:usmco} applies this greedily, masking the lowest-scoring skill and stopping once no score falls below a threshold $\tau$. Only unlabeled queries and their raw contexts are required. Rebuilding the memory dominates the cost at $O(K^2)$ rebuilds, 6--10 minutes per mask step in our setup and negligible beside training.

%======================================================================
\section{Experiments}
\label{sec:exp}
%======================================================================

In this section we evaluate both interventions in detail. Matched-protocol comparisons against strong self-evolving baselines establish their effectiveness, and ablations isolate each design choice.

\subsection{Setup}
Our base self-evolving agent framework is \textsc{MemSkill} \citep{zhang2026memskill}, whose skill-extraction policy agent is Llama-3.3-70B-Instruct \citep{huang2024llama}, matching the original; all experiments run on 8$\times$H20 GPUs. We reproduce the baseline strictly under its original settings (Table~\ref{tab:main}). Our methods are compared under a fully matched protocol: \CASS{} differs from \textsc{MemSkill} only in the gate, and \uSMCO{} is applied only before inference. Evaluation spans four domains: LoCoMo \citep{maharana2024evaluating}, LongMemEval \citep{wu2025longmemeval}, HotpotQA \citep{yang2018hotpotqa} and ALFWorld \citep{shridhar2020alfworld}. Unless noted, we report mean $\pm$ std over three seeds under a fixed judge.

\subsection{Main Results}

We compare our method against state-of-the-art agents: \textsc{CoN} \citep{yu2024chain}, \textsc{ReadAgent} \citep{lee2024human}, \textsc{MemoryBank} \citep{zhong2024memorybank}, \textsc{A-MEM} \citep{xu2026mem}, \textsc{Mem0} \citep{chhikara2025mem0}, \textsc{MemoryOS} \citep{kang2025memory} and \textsc{MemSkill} \citep{zhang2026memskill}; see Table~\ref{tab:main}. Our method delivers strong performance across the three benchmarks and consistently improves on the baseline. We attribute this to what \CASS{} changes in \textsc{MemSkill}'s gating mechanism: because a candidate is admitted only when its contribution survives the coalition audit, the skills that accumulate are mechanistically more reliable and cooperate at the coalition level with those already in the bank.

\subsection{\CASS{} Experiments}

\begin{figure}[t]
\centering
\begin{tikzpicture}[
  cell/.style={minimum width=4.2mm,minimum height=3.3mm,inner sep=0pt,font=\scriptsize},
  Ac/.style={cell,draw=green!55!black,fill=green!30,rounded corners=1pt,font=\scriptsize\bfseries},
  Rj/.style={cell,draw=red!55!black,fill=red!18,rounded corners=1pt},
  lab/.style={font=\scriptsize},
  ttl/.style={font=\scriptsize\bfseries},
]
\node[ttl] at (1.35,0.78) {\CASS{}};
\node[lab] at (0.30,0.40) {13};
\node[lab] at (0.76,0.40) {14};
\node[lab] at (1.22,0.40) {15};
\node[lab] at (1.68,0.40) {16};
\node[lab] at (2.22,0.40) {$K$};
\node[lab,anchor=east] at (0.08,0.00) {s1};
\node[Rj] at (0.30,0.00) {R};
\node[Rj] at (0.76,0.00) {R};
\node[Rj] at (1.22,0.00) {R};
\node[Rj] at (1.68,0.00) {R};
\node[lab,anchor=west] at (2.02,0.00) {$10\!\to\!10$};
\node[lab,anchor=east] at (0.08,-0.38) {s2};
\node[Rj] at (0.30,-0.38) {R};
\node[Rj] at (0.76,-0.38) {R};
\node[Ac] at (1.22,-0.38) {A};
\node[Rj] at (1.68,-0.38) {R};
\node[lab,anchor=west] at (2.02,-0.38) {$10\!\to\!11$};
\node[lab,anchor=east] at (0.08,-0.76) {s3};
\node[Rj] at (0.30,-0.76) {R};
\node[Rj] at (0.76,-0.76) {R};
\node[Rj] at (1.22,-0.76) {R};
\node[Rj] at (1.68,-0.76) {R};
\node[lab,anchor=west] at (2.02,-0.76) {$10\!\to\!10$};
\node[ttl] at (5.20,0.78) {\textsc{MemSkill}};
\node[lab] at (4.15,0.40) {13};
\node[lab] at (4.61,0.40) {14};
\node[lab] at (5.07,0.40) {15};
\node[lab] at (5.53,0.40) {16};
\node[lab] at (6.07,0.40) {$K$};
\node[lab,anchor=east] at (3.93,0.00) {s1};
\node[Ac] at (4.15,0.00) {A};
\node[Rj] at (4.61,0.00) {R};
\node[Ac] at (5.07,0.00) {A};
\node[Ac] at (5.53,0.00) {A};
\node[lab,anchor=west] at (5.87,0.00) {$10\!\to\!13$};
\node[lab,anchor=east] at (3.93,-0.38) {s2};
\node[Ac] at (4.15,-0.38) {A};
\node[Ac] at (4.61,-0.38) {A};
\node[Rj] at (5.07,-0.38) {R};
\node[Rj] at (5.53,-0.38) {R};
\node[lab,anchor=west] at (5.87,-0.38) {$10\!\to\!12$};
\node[lab,anchor=east] at (3.93,-0.76) {s3};
\node[Rj] at (4.15,-0.76) {R};
\node[Rj] at (4.61,-0.76) {R};
\node[Ac] at (5.07,-0.76) {A};
\node[Rj] at (5.53,-0.76) {R};
\node[lab,anchor=west] at (5.87,-0.76) {$10\!\to\!11$};
\draw[draw=black!30] (3.42,0.95) -- (3.42,-0.95);
\end{tikzpicture}
\caption{Gate decisions at the four outer epochs, and the resulting bank size. \CASS{} accepts $1/12$, \textsc{MemSkill} $6/12$.}
\label{fig:accept}
\end{figure}

\begin{table}[t]
\centering
\scriptsize
\setlength{\tabcolsep}{3pt}
\begin{tabular}{lrrr}
\toprule
Benchmark & \textsc{MemSkill} $e_{12}$ & \textsc{MemSkill} final & $\Delta$ \\
\midrule
LoCoMo & 0.5653 & 0.5303 & $-3.50$ \\
LME & 0.5000 & 0.5248 & $+2.48$ \\
HP-50 & 0.3008 & 0.2695 & $-3.13$ \\
\midrule
Avg. & 0.4554 & 0.4415 & $-1.38$ \\
\bottomrule
\end{tabular}
\caption{$K$-controlled ablation of the gate's additions.}
\label{tab:kcontrol}
\end{table}

Across 12 candidate-skill proposals (4 gate decisions per seed, 3 seeds), \CASS{} accepts 1 while \textsc{MemSkill} accepts 6 (Figure~\ref{fig:accept}). A low acceptance rate is a virtue only if what it rejects deserved rejecting, so we audit what the outcome gate admitted. Holding $K$ fixed, we compare the bank at $e_{12}$ against the trained bank containing those additions (Table~\ref{tab:kcontrol}): they are collectively net-negative ($-1.38$~pp on average), although every one satisfied the outcome criterion when admitted, which is the endpoint ambiguity of Theorem~\ref{thm:existence} in practice. Figure~\ref{fig:case} shows what the two gates admit.

\begin{figure}[t]
\centering
\begin{skillbox}{casered}{Capture Temporal Context \textnormal{--- accepted by the outcome gate}}
\scriptsize
\textbf{Purpose:} Capture temporal context and relationships between events from the text chunk.\\[1pt]
\textbf{When to use:} The text chunk mentions specific dates, time frames, or events related to each other.\\[1pt]
\textbf{How to apply:} Attribute the temporal context to the correct events; ensure it is specific, relevant, and includes any causal relationships.
\end{skillbox}
\begin{skillbox}{casegreen}{Capture Entity Attributes \textnormal{--- accepted by \CASS{}}}
\scriptsize
\textbf{Purpose:} Capture detailed, factual information about entities from the text chunk, including attributes, relationships, and preferences.\\[1pt]
\textbf{When to use:} The text chunk mentions specific details about an entity's attributes, relationships, or preferences.\\[1pt]
\textbf{How to apply:} Attribute the detail to the correct entity; ensure the detail is specific, relevant, and factual.
\end{skillbox}
\begin{skillbox}{figpurple}{Insert New Memory \textnormal{--- masked by \uSMCO{} at the target}}
\scriptsize
\textbf{Purpose:} Capture new, durable facts from the current text chunk that are missing in memory.\\[1pt]
\textbf{When to use:} The text chunk introduces new facts, events, plans, or context worth storing.\\[1pt]
\textbf{How to apply:} Compare against retrieved memories to avoid duplicates; split distinct facts into separate items.
\end{skillbox}
\setlength{\abovecaptionskip}{3pt}
\caption{\textbf{Case study.} A coalition-polluting skill accepted by the baseline gate, a coalition-aware skill accepted by \CASS{}, and a utility-reversing skill masked by \uSMCO{}.}
\label{fig:case}
\end{figure}

\subsection{\uSMCO{} Experiments}
\label{ssec:usmco-exp}

\begin{table}[t]
\vspace{-7pt}
\centering
\scriptsize
\begin{tabular*}{\columnwidth}{@{\extracolsep{\fill}} l l c c @{}}
\toprule
Bank & $K$ & Masked skills & $\Delta$ HP-50 (pp) \\
\midrule
\CASS{} s1 & 10$\to$9 & \texttt{AP} & $2.13$ \\
\CASS{} s2 & 11$\to$9 & \texttt{IN}, \texttt{UP} & $4.09$ \\
\CASS{} s3 & 10$\to$8 & \texttt{ES}, \texttt{IN} & $4.55$ \\
\midrule
\textsc{MemSkill} s1 & 13$\to$9 & \texttt{ES}, \texttt{IN}, \texttt{TD}, \texttt{AP} & $5.01$ \\
\textsc{MemSkill} s2 & 12$\to$10 & \texttt{IN}, \texttt{ES} & $8.58$ \\
\textsc{MemSkill} s3 & 11$\to$9 & \texttt{TD}, \texttt{IN} & $7.92$ \\
\bottomrule
\end{tabular*}
\caption{\uSMCO{} on the six trained banks with an unlabeled 20-query target probe; $\Delta$ HP-50 is the LLM-Judge change from masking. Skill codes follow Table~\ref{tab:audit}.}
\vspace{-6pt}
\label{tab:usmco-3seed}
\end{table}

Table~\ref{tab:usmco-3seed} reports \uSMCO{} on all six trained banks. \texttt{Skill\_IN} is masked on 5/6 of them with no target label: the skill the audit finds carrying the bank at the source (Table~\ref{tab:audit}) is the one the target probe rejects, which is cross-domain utility reversal detected label-free, and Figure~\ref{fig:case} shows the skill itself. Masking improves every bank, but it improves the \textsc{MemSkill} banks twice as much as the \CASS{} banks ($7.17$ vs.\ $3.59$~pp on average): the pollution \CASS{} keeps out at the gate is what \uSMCO{} must remove afterwards.

\subsection{More Discussion}
\label{sec:disc}

\paragraph{Reliability is a coalition-level property.} Coalition pollution and cross-domain utility reversal share one root: reliability is a joint property of skill, bank and distribution, not an attribute of the skill. This is why judging a skill in isolation fails. The margins an outcome gate acts on lie below the resolution of a single-skill audit (Figure~\ref{fig:reliability-trajectories}a), and Theorem~\ref{thm:existence} shows the limit is structural rather than statistical: no amount of outcome sampling separates a bank that improved from one that only appeared to. \CASS{} therefore scores coalitions, not skills.

\paragraph{Why retrieval quality.} At deployment the same question returns without labels, and the obvious signal is the agent's own preference. It fails. The drop in selection-policy entropy ranks \texttt{Skill\_IN} as the bank's \emph{most preferred} skill, precisely the one the target rejects, and anti-correlates with the label-based ranking ($\rho=-0.18$, Table~5); that policy was trained under the outcome gate and inherits its bias. Retrieval quality is measured on the target itself and is decoupled from the training reward, reaching $\rho=+0.76$.

\paragraph{Cost.} \CASS{} adds eight coalition evaluations per outer epoch, about $5\%$ of a 20-hour run, and \uSMCO{} runs once before deployment at $O(K^2)$ memory rebuilds. Neither adds any inference-time cost.

%======================================================================
\section{Conclusion}
%======================================================================

We recast skill reliability as a coalition-level property: a skill's contribution depends on the bank around it and on the domain it is deployed in. Under this view, outcome-only gates are provably blind to the two failures we identify, coalition pollution and cross-domain utility reversal. \CASS{} addresses the first during training by scoring sampled coalitions rather than bank-level outcomes, and \uSMCO{} addresses the second after transfer using only unlabeled target queries. Across three seeds both improve the reliability-critical regimes they target. Two directions follow: adapting the coalition budget to how close candidate banks are, and extending the coalition view beyond textual skills.

%======================================================================
\bibliography{cass_paper}

@article{xu2026agent,
  title={Agent skills for large language models: Architecture, acquisition, security, and the path forward},
  author={Xu, Renjun and Yan, Yang},
  journal={arXiv preprint arXiv:2602.12430},
  year={2026}
}

@article{xia2026skillrl,
  title={Skillrl: Evolving agents via recursive skill-augmented reinforcement learning},
  author={Xia, Peng and Chen, Jianwen and Wang, Hanyang and Liu, Jiaqi and Zeng, Kaide and Wang, Yu and Han, Siwei and Zhou, Yiyang and Zhao, Xujiang and Chen, Haifeng and others},
  journal={arXiv preprint arXiv:2602.08234},
  year={2026}
}

@article{zhang2026memskill,
  title={MemSkill: Learning and Evolving Memory Skills for Self-Evolving Agents},
  author={Zhang, Haozhen and Long, Quanyu and Bao, Jianzhu and Feng, Tao and Zhang, Weizhi and Yue, Haodong and Wang, Wenya},
  journal={arXiv preprint arXiv:2602.02474},
  year={2026}
}

@article{yang2026skillopt,
  title={Skillopt: Executive strategy for self-evolving agent skills},
  author={Yang, Yifan and Gong, Ziyang and Huang, Weiquan and Yang, Qihao and Zhou, Ziwei and Huang, Zisu and Li, Yan and Gao, Xuemei and Dai, Qi and Liu, Bei and others},
  journal={arXiv preprint arXiv:2605.23904},
  year={2026}
}

@article{shi2026skill1,
  title={Skill1: Unified evolution of skill-augmented agents via reinforcement learning},
  author={Shi, Yaorui and Chen, Yuxin and Lu, Zhengxi and Miao, Yuchun and Liu, Shugui and Gu, Qi and Cai, Xunliang and Wang, Xiang and Zhang, An},
  journal={arXiv preprint arXiv:2605.06130},
  year={2026}
}

@article{ouyang2026skillos,
  title={Skillos: Learning skill curation for self-evolving agents},
  author={Ouyang, Siru and Yan, Jun and Chen, Yanfei and Han, Rujun and Wang, Zifeng and Mishra, Bhavana Dalvi and Meng, Rui and Li, Chun-Liang and Jiao, Yizhu and Zha, Kaiwen and others},
  journal={arXiv preprint arXiv:2605.06614},
  year={2026}
}

@inproceedings{
berthon2026skill,
title={Skill Neologisms: Towards Skill-based Continual Learning},
author={Antonin Berthon and Nicol{\'a}s Astorga and Mihaela van der Schaar},
booktitle={Forty-third International Conference on Machine Learning},
year={2026},
}

@article{li2026skillsbench,
  title={SkillsBench: Benchmarking how well agent skills work across diverse tasks},
  author={Li, Xiangyi and Liu, Yimin and Chen, Wenbo and You, Bingran and Di, Zonglin and He, Yifeng and Zheng, Shenghan and Choe, Kyoung Whan and Sun, Jiankai and Wang, Shuyi and others},
  journal={arXiv preprint arXiv:2602.12670},
  year={2026}
}

@article{huang2026raw,
  title={From raw experience to skill consumption: A systematic study of model-generated agent skills},
  author={Huang, Zisu and Xu, Jingwen and Yang, Yifan and Gong, Ziyang and Yang, Qihao and Tian, Muzhao and Wang, Xiaohua and Lv, Changze and Gao, Xuemei and Dai, Qi and others},
  journal={arXiv preprint arXiv:2605.23899},
  year={2026}
}

@article{qiu2026autorefine,
  title={AutoRefine: From Trajectories to Reusable Expertise for Continual LLM Agent Refinement},
  author={Qiu, Libin and Gao, Zhirong and Chen, Junfu and Ye, Yuhang and Huang, Weizhi and Xue, Xiaobo and Qiu, Wenkai and Tang, Shuo},
  journal={arXiv preprint arXiv:2601.22758},
  year={2026}
}

@article{ni2026trace2skill,
  title={Trace2skill: Distill trajectory-local lessons into transferable agent skills},
  author={Ni, Jingwei and Liu, Yihao and Liu, Xinpeng and Sun, Yutao and Zhou, Mengyu and Cheng, Pengyu and Wang, Dexin and Zhao, Erchao and Jiang, Xiaoxi and Jiang, Guanjun},
  journal={arXiv preprint arXiv:2603.25158},
  year={2026}
}

@article{zhang2026coevoskills,
  title={Coevoskills: Self-evolving agent skills via co-evolutionary verification},
  author={Zhang, Hanrong and Fan, Shicheng and Zou, Henry Peng and Chen, Yankai and Wang, Zhenting and Zhou, Jiayu and Li, Chengze and Huang, Wei-Chieh and Yao, Yifei and Zheng, Kening and others},
  journal={arXiv preprint arXiv:2604.01687},
  year={2026}
}

@article{chen2026skillcraft,
  title={Skillcraft: Can LLM agents learn to use tools skillfully?},
  author={Chen, Shiqi and Gai, Jingze and Zhou, Ruochen and Zhang, Jinghan and Zhu, Tongyao and Li, Junlong and Wang, Kangrui and Wang, Zihan and Chen, Zhengyu and Kaleb, Klara and others},
  journal={arXiv preprint arXiv:2603.00718},
  year={2026}
}

@article{li2026skillgraph,
  title={SkillGraph: Skill-Augmented Reinforcement Learning for Agents via Evolving Skill Graphs},
  author={Li, Xiaoyuan and Li, Moxin and Bao, Keqin and Ma, Yubo and Wang, Wenjie and Liu, Dayiheng and Feng, Fuli},
  journal={arXiv preprint arXiv:2605.12039},
  year={2026}
}

@article{su2026skill,
  title={Skill retrieval augmentation for agentic ai},
  author={Su, Weihang and Long, Jianming and Ai, Qingyao and He, Qiaozhi and Tang, Yichen and Wang, Changyue and Tu, Yiteng and Wang, Yingbo and Liu, Yiqun},
  journal={arXiv preprint arXiv:2604.24594},
  year={2026}
}

@article{shen2026skillfoundry,
  title={Skillfoundry: Building self-evolving agent skill libraries from heterogeneous scientific resources},
  author={Shen, Shuaike and Cheng, Wenduo and Ma, Mingqian and Turcan, Alistair and Zhang, Martin Jinye and Ma, Jian},
  journal={arXiv preprint arXiv:2604.03964},
  year={2026}
}

@article{wei2026agentic,
  title={Agentic reasoning for large language models},
  author={Wei, Tianxin and Li, Ting-Wei and Liu, Zhining and Ning, Xuying and Yang, Ze and Zou, Jiaru and Zeng, Zhichen and Qiu, Ruizhong and Lin, Xiao and Fu, Dongqi and others},
  journal={arXiv preprint arXiv:2601.12538},
  year={2026}
}

@article{wang2023voyager,
  title={Voyager: An open-ended embodied agent with large language models},
  author={Wang, Guanzhi and Xie, Yuqi and Jiang, Yunfan and Mandlekar, Ajay and Xiao, Chaowei and Zhu, Yuke and Fan, Linxi and Anandkumar, Anima},
  journal={arXiv preprint arXiv:2305.16291},
  year={2023}
}

@article{agarwal2025gpt,
  title={gpt-oss-120b \& gpt-oss-20b model card},
  author={Agarwal, Sandhini and Ahmad, Lama and Ai, Jason and Altman, Sam and Applebaum, Andy and Arbus, Edwin and Arora, Rahul K and Bai, Yu and Baker, Bowen and Bao, Haiming and others},
  journal={arXiv preprint arXiv:2508.10925},
  year={2025}
}

@article{yang2025qwen3,
  title={Qwen3 technical report},
  author={Yang, An and Li, Anfeng and Yang, Baosong and Zhang, Beichen and Hui, Binyuan and Zheng, Bo and Yu, Bowen and Gao, Chang and Huang, Chengen and Lv, Chenxu and others},
  journal={arXiv preprint arXiv:2505.09388},
  year={2025}
}

@article{huang2024llama,
  title={The llama 3 herd of models},
  author={Huang, Kunal Chawla and Lakhotia, Kushal and Huang, Kyle and Chen, Lailin and Garg, Lakshya and Lavender, A and Silva, Leandro and Bell, Lee and Zhang, Lei and Guo, Liangpeng and others},
  journal={preprint},
  year={2024}
}

@article{chen2026learning,
  title={Learning to self-verify makes language models better reasoners},
  author={Chen, Yuxin and Wang, Yu and Zhang, Yi and Ye, Ziang and Cai, Zhengzhou and Shi, Yaorui and Gu, Qi and Su, Hui and Cai, Xunliang and Wang, Xiang and others},
  journal={arXiv preprint arXiv:2602.07594},
  year={2026}
}

@article{yu2026dapo,
  title={Dapo: An open-source llm reinforcement learning system at scale},
  author={Yu, Qiying and Zhang, Zheng and Zhu, Ruofei and Yuan, Yufeng and Zuo, Xiaochen and Yue, Yu and Dai, Weinan and Fan, Tiantian and Liu, Gaohong and Liu, Lingjun and others},
  journal={Advances in Neural Information Processing Systems},
  volume={38},
  pages={113222--113244},
  year={2026}
}

@inproceedings{maharana2024evaluating,
  title={Evaluating very long-term conversational memory of llm agents},
  author={Maharana, Adyasha and Lee, Dong-Ho and Tulyakov, Sergey and Bansal, Mohit and Barbieri, Francesco and Fang, Yuwei},
  booktitle={Proceedings of the 62nd Annual Meeting of the Association for Computational Linguistics (Volume 1: Long Papers)},
  pages={13851--13870},
  year={2024}
}

@inproceedings{
wu2025longmemeval,
title={LongMemEval: Benchmarking Chat Assistants on Long-Term Interactive Memory},
author={Di Wu and Hongwei Wang and Wenhao Yu and Yuwei Zhang and Kai-Wei Chang and Dong Yu},
booktitle={The Thirteenth International Conference on Learning Representations},
year={2025},
}

@inproceedings{yang2018hotpotqa,
  title={HotpotQA: A dataset for diverse, explainable multi-hop question answering},
  author={Yang, Zhilin and Qi, Peng and Zhang, Saizheng and Bengio, Yoshua and Cohen, William and Salakhutdinov, Ruslan and Manning, Christopher D},
  booktitle={Proceedings of the 2018 conference on empirical methods in natural language processing},
  pages={2369--2380},
  year={2018}
}

@article{shridhar2020alfworld,
  title={Alfworld: Aligning text and embodied environments for interactive learning},
  author={Shridhar, Mohit and Yuan, Xingdi and C{\^o}t{\'e}, Marc-Alexandre and Bisk, Yonatan and Trischler, Adam and Hausknecht, Matthew},
  journal={arXiv preprint arXiv:2010.03768},
  year={2020}
}

@article{wu2023autogen,
  title={Autogen: Enabling next-gen llm applications via multi-agent conversation},
  author={Wu, Qingyun and Bansal, Gagan and Zhang, Jieyu and Wu, Yiran and Li, Beibin and Zhu, Erkang and Jiang, Li and Zhang, Xiaoyun and Zhang, Shaokun and Liu, Jiale and others},
  journal={arXiv preprint arXiv:2308.08155},
  year={2023}
}

@article{xu2026mem,
  title={A-mem: Agentic memory for llm agents},
  author={Xu, Wujiang and Liang, Zujie and Mei, Kai and Gao, Hang and Tan, Juntao and Zhang, Yongfeng},
  journal={Advances in Neural Information Processing Systems},
  volume={38},
  pages={17577--17604},
  year={2026}
}

@article{wang2025mirix,
  title={Mirix: Multi-agent memory system for llm-based agents},
  author={Wang, Yu and Chen, Xi},
  journal={arXiv preprint arXiv:2507.07957},
  year={2025}
}

@inproceedings{zhao2024expel,
  title={Expel: Llm agents are experiential learners},
  author={Zhao, Andrew and Huang, Daniel and Xu, Quentin and Lin, Matthieu and Liu, Yong-Jin and Huang, Gao},
  booktitle={Proceedings of the AAAI Conference on Artificial Intelligence},
  volume={38},
  number={17},
  pages={19632--19642},
  year={2024}
}

@article{chhikara2025mem0,
  title={Mem0: Building production-ready ai agents with scalable long-term memory},
  author={Chhikara, Prateek and Khant, Dev and Aryan, Saket and Singh, Taranjeet and Yadav, Deshraj},
  journal={arXiv preprint arXiv:2504.19413},
  year={2025}
}

@article{zhang2025memgen,
  title={Memgen: Weaving generative latent memory for self-evolving agents},
  author={Zhang, Guibin and Fu, Muxin and Yan, Shuicheng},
  journal={arXiv preprint arXiv:2509.24704},
  year={2025}
}

@article{ouyang2025reasoningbank,
  title={Reasoningbank: Scaling agent self-evolving with reasoning memory},
  author={Ouyang, Siru and Yan, Jun and Hsu, I and Chen, Yanfei and Jiang, Ke and Wang, Zifeng and Han, Rujun and Le, Long T and Daruki, Samira and Tang, Xiangru and others},
  journal={arXiv preprint arXiv:2509.25140},
  year={2025}
}

@article{wei2025evo,
  title={Evo-memory: Benchmarking llm agent test-time learning with self-evolving memory},
  author={Wei, Tianxin and Sachdeva, Noveen and Coleman, Benjamin and He, Zhankui and Bei, Yuanchen and Ning, Xuying and Ai, Mengting and Li, Yunzhe and He, Jingrui and Chi, Ed H and others},
  journal={arXiv preprint arXiv:2511.20857},
  year={2025}
}

@article{li2026organizing,
  title={Organizing, orchestrating, and benchmarking agent skills at ecosystem scale},
  author={Li, Hao and Mu, Chunjiang and Chen, Jianhao and Ren, Siyue and Cui, Zhiyao and Zhang, Yiqun and Bai, Lei and Hu, Shuyue},
  journal={arXiv preprint arXiv:2603.02176},
  year={2026}
}

@article{he2026openclaw,
  title={Openclaw as language infrastructure: A case-centered survey of a public agent ecosystem in the wild},
  author={He, Chaoyue and Zhou, Xin and Wang, Di and Xu, Hong and Liu, Wei and Miao, Chunyan},
  year={2026},
  publisher={Preprints}
}

@article{liang2026skillnet,
  title={Skillnet: Create, evaluate, and connect ai skills},
  author={Liang, Yuan and Zhong, Ruobin and Xu, Haoming and Jiang, Chen and Zhong, Yi and Fang, Runnan and Gu, Jia-Chen and Deng, Shumin and Yao, Yunzhi and Wang, Mengru and others},
  journal={arXiv preprint arXiv:2603.04448},
  year={2026}
}

@article{mi2026procmem,
  title={ProcMEM: Learning Reusable Procedural Memory from Experience via Non-Parametric PPO for LLM Agents},
  author={Mi, Qirui and Ma, Zhijian and Yang, Mengyue and Li, Haoxuan and Wang, Yisen and Zhang, Haifeng and Wang, Jun},
  journal={arXiv preprint arXiv:2602.01869},
  year={2026}
}

@article{alzubi2026evoskill,
  title={Evoskill: Automated skill discovery for multi-agent systems},
  author={Alzubi, Salaheddin and Provenzano, Noah and Bingham, Jaydon and Chen, Weiyuan and Vu, Tu},
  journal={arXiv preprint arXiv:2603.02766},
  year={2026}
}

@inproceedings{fang2026memp,
  title={Memp: Exploring agent procedural memory},
  author={Fang, Runnan and Liang, Yuan and Wang, Xiaobin and Wu, Jialong and Qiao, Shuofei and Xie, Pengjun and Huang, Fei and Chen, Huajun and Zhang, Ningyu},
  booktitle={Findings of the Association for Computational Linguistics: ACL 2026},
  pages={17490--17502},
  year={2026}
}

@inproceedings{wang2026reinforcement,
  title={Reinforcement learning for self-improving agent with skill library},
  author={Wang, Jiongxiao and Yan, Qiaojing and Wang, Yawei and Tian, Yijun and Mishra, Soumya Smruti and Xu, Zhichao and Gandhi, Megha and Xu, Panpan and Cheong, Lin Lee},
  booktitle={Proceedings of the 64th Annual Meeting of the Association for Computational Linguistics (Volume 1: Long Papers)},
  pages={1529--1550},
  year={2026}
}

@article{lu2026skill0,
  title={Skill0: In-context agentic reinforcement learning for skill internalization},
  author={Lu, Zhengxi and Yao, Zhiyuan and Wu, Jinyang and Han, Chengcheng and Gu, Qi and Cai, Xunliang and Lu, Weiming and Xiao, Jun and Zhuang, Yueting and Shen, Yongliang},
  journal={arXiv preprint arXiv:2604.02268},
  year={2026}
}

@article{vishe2026skill,
  title={Skill-R1: Agent skill evolution via reinforcement learning},
  author={Vishe, Yash and Surana, Rohan and Jiang, Xunyi and Huang, Zihan and Li, Xintong and Kuang, Nikki Lijing and Yu, Tong and Rossi, Ryan A and Shang, Jingbo and McAuley, Julian and others},
  journal={arXiv preprint arXiv:2605.09359},
  year={2026}
}

@article{zhu2026skill0,
  title={Skill0. 5: Joint Skill Internalization and Utilization for Out-of-Distribution Generalization in Agentic Reinforcement Learning},
  author={Zhu, Jiapeng and Yu, Jianxiang and Zhao, Yibo and Han, Chengcheng and Gu, Qi and Cai, Xunliang and Li, Xiang and Qian, Weining},
  journal={arXiv preprint arXiv:2605.28424},
  year={2026}
}

@article{shen2026dynamic,
  title={Dynamic skill lifecycle management for agentic reinforcement learning},
  author={Shen, Junhao and Zhang, Teng and Zhao, Xiaoyan and Cheng, Hong},
  journal={arXiv preprint arXiv:2605.10923},
  year={2026}
}

@article{tu2026dynamic,
  title={Dynamic Dual-Granularity Skill Bank for Agentic RL},
  author={Tu, Songjun and Xu, Chengdong and Zhang, Qichao and Zhang, Yaocheng and Lan, Xiangyuan and Li, Linjing and Li, Dong and Zhao, Dongbin},
  journal={arXiv preprint arXiv:2603.28716},
  year={2026}
}

@article{xu2026skill,
  title={Skill Reuse as Compression in Agentic RL},
  author={Xu, Zhikun and Feng, Yu and Dineen, Jacob and Shi, Taiwei and Zhao, Jieyu and Zhou, Ben},
  journal={arXiv preprint arXiv:2605.31509},
  year={2026}
}

@incollection{shapley1953value,
  title={A value for n-person games},
  author={Shapley, Lloyd S and others},
  booktitle={Contributions to the Theory of Games},
  year={1953},
  publisher={Princeton University Press Princeton}
}

@article{lundberg2017unified,
  title={A unified approach to interpreting model predictions},
  author={Lundberg, Scott M and Lee, Su-In},
  journal={Advances in neural information processing systems},
  volume={30},
  year={2017}
}

@inproceedings{sundararajan2020many,
  title={The many Shapley values for model explanation},
  author={Sundararajan, Mukund and Najmi, Amir},
  booktitle={International conference on machine learning},
  pages={9269--9278},
  year={2020},
  organization={PMLR}
}

@book{grabisch2016set,
  title={Set functions, games and capacities in decision making},
  author={Grabisch, Michel and others},
  volume={46},
  year={2016},
  publisher={Springer}
}

@article{wang2020tent,
  title={Tent: Fully test-time adaptation by entropy minimization},
  author={Wang, Dequan and Shelhamer, Evan and Liu, Shaoteng and Olshausen, Bruno and Darrell, Trevor},
  journal={arXiv preprint arXiv:2006.10726},
  year={2020}
}

@article{iwasawa2021test,
  title={Test-time classifier adjustment module for model-agnostic domain generalization},
  author={Iwasawa, Yusuke and Matsuo, Yutaka},
  journal={Advances in Neural Information Processing Systems},
  volume={34},
  pages={2427--2440},
  year={2021}
}

@inproceedings{kocsis2006bandit,
  title={Bandit based monte-carlo planning},
  author={Kocsis, Levente and Szepesv{\'a}ri, Csaba},
  booktitle={European conference on machine learning},
  pages={282--293},
  year={2006},
  organization={Springer}
}

@inproceedings{yu2024chain,
  title={Chain-of-note: Enhancing robustness in retrieval-augmented language models},
  author={Yu, Wenhao and Zhang, Hongming and Pan, Xiaoman and Cao, Peixin and Ma, Kaixin and Li, Jian and Wang, Hongwei and Yu, Dong},
  booktitle={Proceedings of the 2024 conference on empirical methods in natural language processing},
  pages={14672--14685},
  year={2024}
}

@inproceedings{lee2024human,
  title={A human-inspired reading agent with gist memory of very long contexts},
  author={Lee, Kuang-Huei and Chen, Xinyun and Furuta, Hiroki and Canny, John and Fischer, Ian},
  booktitle={Proceedings of the 41st International Conference on Machine Learning},
  pages={26396--26415},
  year={2024}
}

@inproceedings{zhong2024memorybank,
  title={Memorybank: Enhancing large language models with long-term memory},
  author={Zhong, Wanjun and Guo, Lianghong and Gao, Qiqi and Ye, He and Wang, Yanlin},
  booktitle={Proceedings of the AAAI conference on artificial intelligence},
  volume={38},
  number={17},
  pages={19724--19731},
  year={2024}
}

@inproceedings{kang2025memory,
  title={Memory os of ai agent},
  author={Kang, Jiazheng and Ji, Mingming and Zhao, Zhe and Bai, Ting},
  booktitle={Proceedings of the 2025 Conference on Empirical Methods in Natural Language Processing},
  pages={25972--25981},
  year={2025}
}

%======================================================================
\end{document}